\documentclass[preprint, 10pt, 3p]{elsarticle}

\usepackage{amssymb}
\usepackage{amsthm}
\usepackage{amsmath}

\newtheorem{definition}{Definition}

\journal{Image and Vision Computing}

\usepackage{xcolor}
\usepackage{algorithm}
\usepackage{algorithmic}
\usepackage{tabularx}
\usepackage{multicol}
\usepackage{cleveref}
\usepackage{tikz}
\usepackage{tikz-cd}
\usepackage{mathtools}
\usepackage{subcaption}
\usepackage{paracol}
\usepackage{titlesec}
\usepackage{url}

\titleformat*{\section}{\large\bfseries}
\titleformat*{\subsection}{\bfseries}

\crefformat{equation}{(#2#1#3)}

\usetikzlibrary{cd, arrows, positioning}
\tikzset{
    shift left/.style={commutative diagrams/shift left={#1}},
    shift right/.style={commutative diagrams/shift right={#1}}
}

\newcommand{\datspace}{\mathcal{X}}
\newcommand{\latspace}{\mathcal{Z}}
\newcommand{\R}{\mathbb{R}}
\newcommand{\N}{\mathbb{N}}
\newcommand{\mean}[2]{\mathbb{E}_{#1}\left[ #2 \right]}
\newcommand{\var}[1]{\mathrm{\mathbf{#1}}}
\newcommand{\norm}[1]{\lVert #1 \rVert}

\DeclarePairedDelimiterX{\infdivx}[2]{(}{)}{%
  #1\;\delimsize\|\;#2%
}
\newcommand{\kldiv}{\mathrm{D}_{KL}\infdivx}

\tikzset{
    pics/space/.style n args={2}{
        code={%
        \draw[thick] (0,1) to [out=10,in=90] (2,0) to [out=270,in=30] (0,-1) to [out=210,in=290]  (-2,0) to [out=110,in=190] (0,1);
        \draw (-1.5,1) node {#1};
        \draw (-0.2,0.2) node {#2};
}},
    U1/.pic={
        \draw[thick, dashed] (0,1) to [out=350,in=90] (2,0) to [out=270,in=30] (1,-1) to [out=210,in=270] (-2,0) to [out=90, in=170] (0,1);
        \draw (-1.2,0.2) node {#1};
    },
    U2/.pic={
        \draw[thick, dashed] (0,2) to [out=330,in=110] (2,0) to [out=290,in=30] (0,-1) to [out=210,in=270] (-2,0) to [out=90, in=150] (0,2);
        \draw (-1,-0.5) node {#1};
    },
    U3/.pic={
        \draw[thick, dashed] (0,1) to [out=0,in=120] (2,0.5) to [out=300,in=30] (1,-1) to [out=210,in=270] (-2,0) to [out=90, in=180] (0,1);
        \draw (1.2,0.3) node {#1};
    },
    Rd/.pic={
        \draw[thick, -] (0,0.3) to (3.2,2);
        \draw[thick, -] (0,0.3) to (7,0);
        \draw[thick, -] (7,0) to (9.9,1.7);
        \draw[thick, -] (3.2,2) to (9.9,1.7);
        \draw (8.5,2.3) node {#1};
    },
    gaussian/.pic={
        \draw[variable=\x, smooth, domain=-3:3, thick] plot ({\x},{1.5*exp(-\x*\x*0.5)});
        \draw (0,0.5) node[align=right,label=below:{#1}] {};
    }
}

\begin{document}

\begin{frontmatter}



\title{Atlas Generative Models\\and Geodesic Interpolation}


\author{Jakob Stolberg-Larsen}

\author{Stefan Sommer}

\address{Department of Computer Science\\University of Copenhagen}

\begin{abstract}
    Generative neural networks have a well recognized ability to estimate underlying manifold structure of high dimensional data. However,
if a single latent space is used, it is not possible to
faithfully represent a manifold with topology different from Euclidean space.
In this work we define the general class of Atlas Generative Models
(AGMs), models with hybrid discrete-continuous latent space that
estimate an atlas on the underlying data manifold together with a
partition of unity on the data space. We identify existing examples of
models from various popular generative paradigms that fit into this
class. Due to the atlas interpretation, ideas from non-linear latent
space analysis and statistics, e.g. geodesic interpolation, which has
previously only been investigated for models with simply connected
latent spaces, may be extended to the entire class of AGMs in a natural
way. We exemplify this by generalizing an algorithm for graph based
geodesic interpolation to the setting of AGMs, and verify its performance
experimentally.
\end{abstract} 

\end{frontmatter}


\section{Introduction}
The ability of deep generative networks to learn complex features of data in an unsupervised fashion has made them a promising tool for dealing with the problem of increasing amounts of unlabelled data and inchoate labelling. A (probabilistic) generating map $G: \latspace \to \datspace$ transforms latent random seeds into synthetic data, usually with $\latspace = \R^d$ and $\datspace=\R^D$ for some $d \ll D$. Via $G$, a low-dimensional manifold structure, which follows high density regions of the data distribution, is learned. 

Among other things, this enables continuous interpolations between points in latent space, rendering in $\datspace$ continuous transformation of samples along the underlying manifold structure of the data distribution. While an obvious option is linear interpolation for Euclidean latent spaces, recent research has also investigated using geodesic interpolations for $\latspace = \R^d$ considered as a Riemannian manifold. The geometric structure is here chosen so that curve length in the latent space matches curve length in the data space for curves restricted to the manifold $G$ \cite{shao_riemannian_2018, chen_metrics_2018, arvanitidis_latent_2018}. The approach yields a more accurate notion of distance and shortest paths in $\latspace$ as it is based on the distance actually traversed in $\datspace$ along the manifold structure.

For inherently non-linear latent spaces, e.g. hybrid discrete-continuous latent space $\latspace \times \mathcal{Y}$ where $\mathcal{Y} = \{1, \ldots, m\}$ for some $m\in \N$, Euclidean distances and linear interpolations are not even well defined. However, in this paper, we suggest that we can still make sense of geodesic interpolations, thus expanding on the geometrical interpretation of deep generative networks. This generalization is important: Even for simple manifolds, a single latent space is not sufficient to accurate represent data spread over the entire manifold.

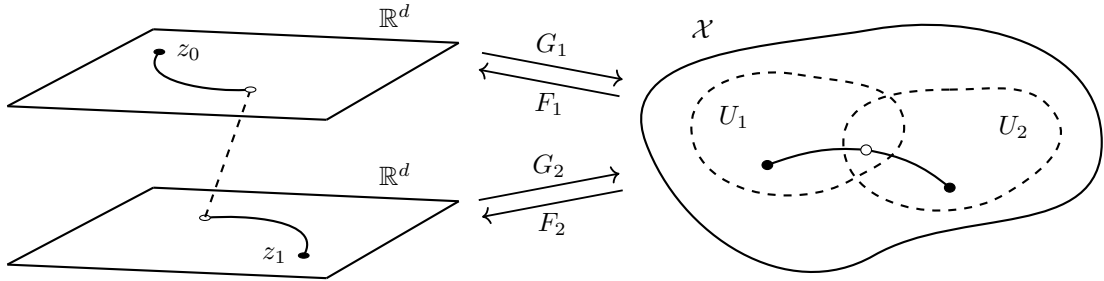
\begin{figure}[t]
    \centering
    \begin{tikzpicture}[auto]
    \draw (5,0) pic[scale=1.5] {space={$\datspace$}{}};
    \draw (4,0.2) pic[scale=0.7] {U1={$U_1$}};
    \draw (6,0) pic[scale=0.7] {U3={$U_2$}};
    \draw (-6.4,0.3) pic[scale=0.6] {Rd={$\R^d$}{}};
    \draw (-6.4,-1.8) pic[scale=0.6] {Rd={$\R^d$}{}};
    \node (a) at (-0.3,1.1) {};
    \node (b) at (1.8,0.7) {};
    \node (c) at (-0.3,-0.9) {};
    \node (d) at (1.8,-0.5) {};
    \path[->, shift left=.75ex, line width=0.7pt]%
    (a) edge node [above] {$G_1$} (b)%
    (b) edge node [below] {$F_1$} (a)%
    (c) edge node [above] {$G_2$} (d)%
    (d) edge node [below] {$F_2$} (c);
    \path   (3.6,-0.3)     coordinate (A)%
            (4.9,-0.1)     coordinate (B)%
            (6,-0.6)       coordinate (C);
            \draw[line width=0.8pt]   (A) to [out=20, in=175] (B)%
            (B) to [out=350, in=140] (C);
            \draw[thin, fill=black] (A) ellipse (2pt and 1.7pt);
            \draw[thin, fill=white] (B) ellipse (2pt and 1.9pt);
            \draw[thin, fill=black] (C) ellipse (2pt and 1.7pt);
    \path   (-4.4, 1.2)     coordinate (D)%
            (-3.2, 0.7)     coordinate (E)%
            (-3.8, -1)      coordinate (F)%
            (-2.5, -1.5)    coordinate (G);
            \draw[line width=0.8pt]   (D) to [out=230, in=185] (E);
            \draw[line width=0.8pt]   (F) to [out=5, in=60] (G);
    \draw[thin, fill=white] (F) ellipse (2pt and 1pt);
    \draw[thin, fill=black] (G) ellipse (2pt and 1pt);
    \draw[dashed, line width=0.8pt]      (E) to (F);
    \draw[thin, fill=black] (D) ellipse (2pt and 1pt);
    \draw[thin, fill=white] (E) ellipse (2pt and 1pt);
    \node at (D) [right = 1mm of D] {$z_0$};
    \node at (G) [left = 1mm of G] {$z_1$};
\end{tikzpicture}
    \caption{Geodesic interpolation for AGMs. Using a partition of unity $\psi : \datspace \to \Delta^{m-1}$ we can detect the areas where charts overlap, and use it to make sense of continuous geodesic interpolation in an otherwise discontinuous latent space.}
    \label{fig:agm}
\end{figure}

The manifold estimating qualities of generative networks, combined with a hybrid discrete-continuous latent space $\latspace \times \mathcal{Y}$, has already lead to an interpretation inspired by the notion of a manifold atlas from differential geometry. For each restriction to some $y \in \mathcal{Y}$, the map $G_y : \latspace \to \datspace$ resembles (the inverse of) a coordinate chart on the immersed manifold (see \cref{fig:agm}). Most notably, it is formally shown in \cite{schonsheck_chart_2020} that multiple charts are actually necessary in order to properly approximate data manifold structure with non-trivial topology. Building on this, we state explicitly some general criteria we expect of a generative model in order for them to fully satisfy the atlas interpretation, and use the terminology \emph{Atlas Generative Models} (AGM) to refer to this class of models. Like with the Chart Auto-Encoder (CAE) of \cite{schonsheck_chart_2020}, these are models that, in addition to chart inverse estimates $G_y$, yield chart estimates $F_y : \datspace \to \latspace$ and a partition of unity $\psi : \datspace \to \Delta^{m-1}$, another concept taken from differential geometry. Here $\Delta^{m-1}$ denotes the standard $(m-1)$-simplex, and $\psi$ may thus through its coordinate functions $\psi_y$ be seen as assigning to each point $x \in \datspace$ the importance of the individual charts in that point, with these summing to one. 

We show how examples of AGMs, in addition to the CAE, are also to be found within the paradigms of Variational Auto-Encoders (VAEs) and Wasserstein Auto-Encoders (WAEs). If one relaxes the requirements by not demanding encoding networks $F_y$, we also see examples from the realm of Generative Adversarial Networks (GANs). Some of the models have grown out of the desire to capture non-trivial topological features of the underlying manifold structure of high-dimensional data, while others have been studied for the purpuses of disentangled representation learning or semi-supervised learning. 

With a concise yet general concept to build upon, we move on to describe a graph based procedure for approximating latent space geodesic paths, generalizing the algorithm from \cite{chen_fast_2019} to the more broad setting of AGMs, thus providing a novel concept of latent space interpolation. Using the partition of unity, we may define areas for which the importance of the charts transcend some threshold, $U_y = \{x\in \datspace \mid \psi_y(x) > \varepsilon\}$ for $\varepsilon \geq 0$. Using the intersections of these, we can identify points in different charts of the latent space in accordance with the manifold structure, and thus make sense of continuously traversing across different charts (see \cref{fig:agm}). As this is coherent with the geometric interpretation of generative networks as manifold estimators, we see it as a constructive first step to expanding the theory of non-linear latent space analysis and statistics, to the class of atlas estimating models.

Our main contributions consist of:
\begin{itemize}
    \item Providing a precise characterization of a class of generative networks, AGMs, which may be thought of as estimating an atlas on the underlying manifold structure of a data distribution, and showing that multiple existing generative models fit into this class. 
    \item Introducing a procedure for approximating latent space geodesic paths for any model in the class of AGMs, thus providing a notion of continuous interpolation novel to this class of models. We note, that even having a notion of continuous interpolation is new, as the trivial example of linear interpolation is not well defined for AGMs, due to the inherently discontinuous nature of their latent space.
    \item Demonstrating empirically that the procedure produces interpolations with comparable qualities to those produced in a non-atlas setting, thus making it a viable tool when working with any AGM.
\end{itemize}

With these contributions, we extend the current ability to interpolate between data points in generative models with single latent spaces to a much wider class of data manifolds: Manifolds that have non-trivial topology and thus need more than a single latent space to be accurately represented. We construct the extension in a general setting to avoid restricting to a specific generative model. Instead, the construction encompasses an entire class of models, the class that we denote AGMs.

We begin the paper with a theory section, in which we briefly present the necessary background for our contributions. In the following section, we present our definition of AGMs and mention the examples of generative models within this class. After that, we present the graph based procedure for geodesic interpolation for AGMs. We end with an experiment section, in which we demonstrate the procedure for a specific AGM trained on the MNIST dataset \cite{deng_mnist_2012}.

\section{Theory}
In this background section, we will first go through the theory and relevant paradigms of generative networks. After this, we will review their relation to the manifold hypothesis, including non-linear latent space analysis, such as geodesic interpolation.

\subsection{Generative Neural Networks}\label{generative-networks}
Most generative neural networks use a (probabilistic) transformation $G : \latspace \to \datspace$ with $\datspace = \R^D$ and $\latspace = \R^d$ for some $d \ll D$. However we will be more concerned with models which have a hybrid discrete-continuous latent space $\latspace \times \mathcal{Y}$, where $\mathcal{Y} = \{1, \ldots, m\}$ for some $m\in \N$, with prior distribution $P(Z,Y) = P(Z)P(Y)$. Learning the generative transformation $G$ is the common aim of generative network models, but there has been quite varying approaches to optimizing the parameters of $G$, ultimately rooted in different theoretical motivations. Let us consider a few:

\textbf{Variational Auto-Encoder (VAE):} In \cite{kingma_auto-encoding_2014} the task of training a generative network is approached by implementing, in addition to a probabilistic decoding network $z \mapsto P(X|z)$, a probabilistic encoding network $x \mapsto Q(Z|x)$, and training these simultaneously to maximize the variational lower bound on the marginal log-likelihood of the data. If the latent space is $\latspace \times \mathcal{Y}$, the encoder is given by $x \mapsto Q(Z|y,x)Q(Y|x)$ and the decoder by $(z,y) \mapsto P(X|z,y)$. The variational lower bound becomes  
\begin{equation}
    \log p(x) \geq -\kldiv{q(y|x)}{p(y)} + \sum_{i=1}^{m}q(y|x) \mathcal{L}(x,y),
    \label{atlas-variational-lower-bound}
\end{equation}
where $\kldiv{\cdot}{\cdot}$ denotes the Kullberg-Leibler divergence between two probability distributions and  
\begin{equation}
    \mathcal{L}(x,y) = -\kldiv{q(z|y,x)}{p(z)} + \mean{z \sim Q(Z|y,x)}{\log p(x|z,y)}
    \label{variational-lower-bound}
\end{equation}
corresponds for a given $y$ to the variational lower bound for a continuous latent space VAE. The significance of the lower bound \cref{atlas-variational-lower-bound} was proven in \cite{kingma_semi-supervised_2014}, where, using it for unlabelled data, it was shown how generative networks can be utilized for semi-supervised classification. It is also the training objective for the InfoCatVAE model of \cite{pineau_infocatvae_2018} which is used to learn disentangled latent representation in a completely unsupervised fashion, by combining the lower bound with a regularizing term for maximizing mutual information inspired by \cite{chen_infogan_2016} (see \cref{infogan-mi}).

\textbf{Generative Adversarial Network (GAN):} Introduced in \cite{goodfellow_generative_2014}, GANs are motivated from a game theoretical perspective. Together with a generating network $G: \latspace \to \datspace$, a discriminative neural network $D: \datspace \to [0,1]$ is implemented, acting as an adversary to the generator. The generator and the discriminator plays a minimax game, which in its equilibrium state minimizes the Jensen-Shannon (JS) divergence between the true data distribution $P_{\mathrm{data}}(X)$ and the distribution of the synthetic data $P_G(X)$. 

To achieve disentangled latent representation the InfoGAN model is introduced in \cite{chen_infogan_2016}. This model has a latent space $\latspace \times \mathcal{Y}$ and adds a discrete variable inference network $x \mapsto Q(Y|x)$, which makes it possible to maximize the mutual information (MI) between the descrete latent variable $y$ and samples $G(z,y)$. The minimax game becomes
\begin{equation}
    \min_{G,Q} \max_D V(G,D) - \lambda L(G,Q),
    \label{min-max-game}
\end{equation}
where $\lambda >0$ is a hyper-parameter and 
\begin{align}
    V(G,D) &= \mean{x \sim P_{\mathrm{data}}(X)}{\log D(x)} + \mean{(z,y) \sim P(Z,Y)}{\log (1 - D(G(z,y)))}, \label{infogan-value} \\
    L(G,Q) &= \mean{y \sim P(Y)}{\mean{x \sim P_G(X|y)}{\log q(y|x)}} + \mathrm{H}[p(y)].
    \label{infogan-mi}
\end{align}
Here $\mathrm{H}[p(y)]$ denotes the differential entropy of the discrete prior $P(Y)$. 

InfoGAN obtains impressively disentangled latent space representation in a completely unsupervised fashion, and it is worth noting that it easily fits into a semi-supervised setting as well. This is done in \cite{spurr_guiding_2017}, which shows how combining the InfoGAN with small amounts of auxiliary label information both increases the quality of synthetic samples, and speeds up convergence of the model.

\textbf{Wasserstein Auto-Encoder (WAE):} The WAE models presented in \cite{tolstikhin_wasserstein_2018} are a flexible class of auto-encoding models, with a particular example being the Adversarial Auto-Encoder (AAE) of \cite{makhzani_adversarial_2016}. They operate with the objective of minimizing the optimal transportation cost between $P_G(X)$ and $P_\mathrm{data}(X)$. If we again assume a latent space $\latspace \times \mathcal{Y}$, the WAE consists of encoding and decoding networks, $F: \datspace \to \latspace \times \mathcal{Y}$ respectively $G : \latspace \times \mathcal{Y} \to \datspace$. These can be probabilistic, but we shall for simplicity assume deterministic encoding and decoding. The optimal transpotation cost is approximated by 
\begin{equation} 
    D_{\mathrm{WAE}}(P_G, P_\mathrm{data}) = \mean{x \sim P_\mathrm{data}(X)}{c(x, G \circ F(x))} + \lambda \mathcal{D}(Q(Z,Y), P(Z,Y)),
    \label{ot-approx}
\end{equation}
where $c : \datspace \times \datspace \to \R_+$ is some measurable cost function, and $\mathcal{D}(Q(Z,Y), P(Z,Y))$ is some divergence measure between $Q(Z,Y)$, which is the image distribution in $\latspace$ of $P_\mathrm{data}(X)$ under the encoder $F$, and the prior distribution $P(Z,Y)$. In \cite{korman_autoencoding_2018}, the Euclidean distance $c = \norm{x - G \circ F(x)}_2$ is used for the reconstruction and the JS-divergence is used as $\mathcal{D}$, as this can be minimized by an adversarial approach. The result is a variation of the WAE-GAN \cite{tolstikhin_wasserstein_2018}, or AAE \cite{makhzani_adversarial_2016}, which also has discrete latent variables. The motivation behind this is to emulate an atlas, enabling the capture of non-trivial homotopical structures of the data distribution, something we elaborate on in \cref{manifold-estimation}.

\textbf{Chart Auto-Encoder (CAE):} The auto-encoding model CAE presented in \cite{schonsheck_chart_2020} is directly based on the notion of a manifold atlas. It has latent space $\latspace \times \mathcal{Y}$ with encoding and decoding maps $F_y : \datspace \to \latspace$ respectively $G_y : \latspace \to \datspace$ for $y = 1, \ldots, m$ and a chart prediction network $P: \datspace \to \Delta^{m-1}$. This is presented with a specific model architecture, well suited for multi-chart representation and a training loss function defined by
\begin{equation}
    L(x) = \Big(\min_{y \in \mathcal{Y}} e_y\Big) - \sum_{y=1}^{m}l_y \log P_y(x),
    \label{cae-loss}
\end{equation}
for $x \in \datspace$, where $e_y = \norm{x - G_y \circ F_y(x)}^2$ and $l_y = \mathrm{softmax}(e_y)$. This model is also topologically and geometrically motivated, i.e. aimed at approximating the underlying manifold structure of the data distribution.
 
\subsection{Manifold Estimation with Generative Networks}\label{manifold-estimation}
In differential geometry, a smooth atlas on a $d$-dimensional manifold $\mathcal{M}$ is a collection of charts $\{\varphi_\alpha : U_\alpha \to \R^d\}$, where $U_\alpha \subset \mathcal{M}$ are open subsets such that $\cup_\alpha U_{\alpha} = \mathcal{M}$. The charts must be homeomorphisms on their image and the transitions $\varphi_\alpha \circ \varphi_\beta^{-1} : \varphi_\beta(U_\alpha \cap U_\beta) \to \R^d$ should be smooth maps. A partition of unity subordinate to an open cover $\{U_\alpha\}$ of $\mathcal{M}$ is a collection of functions $\psi_\alpha : \mathcal{M} \to [0,1]$, which satisfy 
\begin{itemize}
    \item The support $\mathrm{supp}\, \psi_\alpha$ is contained in $U_\alpha$ for all $\alpha$.
    \item Every point $x \in \mathcal{M}$ has a neighbourhood $V$, such that $\mathrm{supp}\, \psi_\alpha \cap V \neq \emptyset$ for only finitely many $\alpha$. 
    \item $\sum_{\alpha}^{} \psi_\alpha(x) = 1$ for all $x \in M$.
\end{itemize}
In particular, a finite partition of unity may be expressed as a map $\psi : \mathcal{M} \to \Delta^{m-1}$, where $\Delta^{m-1}$ is the standard $(m-1)$-simplex, for which the coordinate functions satisfy $\mathrm{supp}\, \psi_i \subset U_i$ for $i = 1, \ldots, m$.

A generative model $G :\latspace \to \datspace$ may be considered a map of manifolds. In particular, $G$ may parametrize an embedded manifold in $\datspace$ if the dimension of $\latspace$ is lower than that of $\datspace$ and it is homeomorphic to its image. As the widely used manifold hypothesis states that high dimensional data is often distributed in proximity of a manifold structure of dimension far lower, it is natural to expect $G$ to estimate this structure.

Unfortunately, there are limitations connected to the choice of a simply connected latent space. In \cite{schonsheck_chart_2020}, it is shown formally that a model with a simply connected subset of $\R^d$ as latent space cannot faithfully represent any manifold structure of non-trivial homotopy type, but it is possible if the latent space consists of a collection of coordinate spaces $\latspace \times \mathcal{Y}$. Having a hybrid discrete-continuous latent space may thus in some scenarios be strictly necessary to obtain approximative qualities of $G$ within a certain margin.  

\subsection{Riemannian Geometry of Generative Networks}\label{riemannian-geometry}
A smooth manifold $\mathcal{M}$ is called a Riemannian manifold, if it is equipped with a so called Riemannian metric. This is the assignment of an inner product $\langle \cdot , \cdot \rangle_{p}$ to the tangent space $T_p \mathcal{M}$ for each point $p \in \mathcal{M}$. The Riemannian metric can be used to define curve lengths of piecewise smooth curves $\gamma : [0,1] \to \mathcal{M}$ as 
\begin{equation}
    L(\gamma) = \int_0^1 \sqrt{\langle \gamma'(t) , \gamma'(t) \rangle_{\gamma(t)}} \, \mathrm{d}t,
    \label{curvelength}
\end{equation}
and in turn gives a notion of distance between points $p_0, p_1 \in \mathcal{M}$, as the infimum $\inf_\gamma L(\gamma)$ taken over all piecewise smooth curves $\gamma$ going from $p_0$ to $p_1$. Geodesic curves on a Riemannian manifold are curves which are locally length minimizing with respect to this distance. 

In case of a map of smooth manifolds, $G : \R^d \to \R^D$, one can get a Riemannian metric on $\R^d$ by taking the pullback of the Euclidean metric in $\R^D$. This is given by assigning to each point $z \in \R^d$ the inner product
\begin{equation}
    \langle u , v \rangle_z = u^\top J_G(z)^\top J_G(z) v
\end{equation}
for all tangent vectors $u,v \in T_z \R^d$, where $J_G(z)$ is the Jacobian matrix of $G$ evaluated in $z$.

The idea of considering the latent space of a generative model $G : \latspace \to \datspace$ with the pullback of the Euclidean metric is presented in \cite{shao_riemannian_2018}, \cite{chen_metrics_2018} and \cite{arvanitidis_latent_2018}, in order to give a more accurate notion of distances and shortest paths in the latent space. As such, linear paths and Euclidean distances in $\latspace$ are replaced with geodesic paths and Riemannian distances. Different approaches to finding geodesic paths with respect to the latent space Riemannian metric are presented in \cite{shao_riemannian_2018}, \cite{chen_metrics_2018} and \cite{arvanitidis_latent_2018}. The Riemannian latent space concept is further expanded with other tools for non-linear latent space analysis and statistics in \cite{kuhnel_latent_2018}. 

To improve efficiency, a graph based approach to approximate geodesics with respect to the Riemannian geometry was presented in \cite{chen_fast_2019}. First a graph is formed in latent space as a k-d tree. To get the nodes for the graph a set of data points $\var{X} = \{x^{(1)}, \ldots, x^{(N)}\}$ are mapped to their respective latent encodings $\var{Z} = \{ z^{(1)}, \ldots, z^{(N)}\}$. Assuming the geodesics are locally close to linear, one proceeds by finding the $k$-nearest neighbours of each node $z^{(i)}$ with respect to the Euclidean distance in $\latspace$, and connecting each of these to $z^{(i)}$ with an edge. These edges are weighted with the length of the interpolation with respect to the Riemannian metric, i.e. using a numerical estimation of \cref{curvelength}. 

For two points $z_0, z_1 \in \latspace$, the geodesic between $z_0$ and $z_1$ is approximated by the shortest path through the graph. More specifically, $z_0$ and $z_1$ are added to the graph and connected with their $k$ nearest neighbours through edges weighted as above. Using A* path search in the graph, the shortest path between $z_0$ and $z_1$ is found. The resulting curve is thus the piecewise linear interpolation in $\latspace$ between nodes along this graph path.

Even though all of the aforementioned research on non-linear latent space analysis build on the manifold interpretation of generative networks, there has, to the best of our knowledge, been no attempts at generalizing the tools presented to a setting where $G : \latspace \times \mathcal{Y} \to \datspace$ estimate an atlas. After laying out exactly which models this includes, by providing a terminological foundation, we shall take a first step in this direction by generalizing the procedure of \cite{chen_fast_2019} to such models.

\section{Atlas Generative Models} \label{theory-agm}
Building on the ideas of \cite{schonsheck_chart_2020} and \cite{korman_autoencoding_2018}, we will now draw up the essential components that make an atlas estimating generative network, which we shall define as the following: 

\begin{definition}\label{agm-definition}
    An Atlas Generative Model (AGM) is a generative model with latent space $\R^d \times \{1, \ldots, m\}$ for some $d,m \in \N$, which post-training yields a family of chart and chart inverse estimates, $F_y: \datspace \to \R^d$ respectively $G_y : \R^d \to \datspace$ for $y=1, \ldots, m$, together with a partition of unity $\psi : \datspace \to \Delta^{m-1}$.

    A generative model which only yields chart inverse estimates $\{G_y\}_{y=1}^m$ and a partition of unity $\psi$, we call a semi-AGM.
\end{definition}
We stress that the encoding and decoding maps $F_y$ respectively $G_y$, only \emph{estimate} charts of an atlas, and thus do not possess all the theoretical properties of a manifold atlas. Most notably, there typically will not be guarantees that $F_y$ are the inverses of $G_y$ on their image, though training objectives of the generative models will often encourage that they are close to that. 

As we shall see, the generative models surveyed in \cref{generative-networks} fall into the class of AGMs or semi-AGMs:

\textbf{CAE:} Designed solely with the purpose of resembling an atlas, the CAE model naturally fits the AGM characterization. Encoding and decoding networks estimate charts $F_y$ and chart inverses $G_y$, and the chart prediction network defines the partition of unity $\psi$.

\textbf{Atlas VAE:} Within the paradigm of VAEs, we can consider the VAE for semi-supervised learning \cite{kingma_semi-supervised_2014} or the InfoCatVAE \cite{pineau_infocatvae_2018}. The discrete inference network defines a partition of unity by letting $\psi_y(x) = q(y|x)$ for all $y \in \mathcal{Y}$. Furthermore we get chart estimates $F_y : \datspace \to \R^d$ and chart inverse estimates $G_y: \R^d \to \datspace$ by letting $F_y(x) = \mean{z \sim Q(Z|y,x)}{z}$ and $G_y(z) = \mean{x \sim P(X|z,y)}{x}$ respectively. These mean values are typically directly available, as encoders and decoders of VAEs usually map a point to the mean value and variance of a multivariate Gaussian distribution. 

These AGMs were initially presented with the purpose of semi-supervised classification, respectively unsupervised, disentangled representation learning. 

\textbf{Atlas WAE:} Unsurprisingly, the WAE model of \cite{korman_autoencoding_2018}, which was directly inspired by the notion of atlases in differential geometry, also falls into the class of AGMs. The encoding and decoding maps directly provide the charts $F_y$ and inverses $G_y$, and the discrete inference network naturally provides a partition of unity like above. While this is a particular example, we note that another cost function $c : \datspace \times \datspace \to \R_+$, as well as divergence term $\mathcal{D}(Q(Z,Y), P(Z,Y))$, could also be used, though we have not seen it explored in other research. 

In cases where encoding or decoding transformations are probabilistic, i.e. $x \mapsto Q(Z,Y|x)$ and $(z,y) \mapsto P(X|z,y)$, chart estimates and inverses are easily obtained by using the mean value of the image distributions, exactly as is the case for Atlas VAEs. 

In \cite{korman_autoencoding_2018}, the Atlas WAE model is used to capture underlying, non-trivial topological structures in the data distribution. We observe in \cref{experiments} how it also produces a disentangled latent representation, which most likely could be further improved by adding an MI-regularizer like in the InfoCatVAE. We are not aware of research into utilizing Atlas WAEs for semi-supervised learning, though this could also be implemented alike \cite{kingma_semi-supervised_2014}. 

\textbf{Atlas GAN:} The adversarial paradigm within generative networks has quickly become one of the most popular. A downside to these models though, is the lack inference networks. This is also partly the case for the InfoGAN model, however the discrete inference network may, as in the previous examples, be used as a partition of unity, which together with the decoding maps $G_y$ makes the InfoGAN a semi-AGM. 

While this category is slightly deficient compared to regular AGMs, we think the significance of the adversarial paradigm makes it worthwile to include. Especially since they share qualities with AGMs, e.g. disentangled representation learning, and since the manifold estimation is also considered a noteworthy quality of GANs. We see in the ss-InfoGAN \cite{spurr_guiding_2017} another example of how charts of (semi-)AGMs may easily be paired with auxiliary semantic labels, in that case improving both sample quality and training efficiency. \\ 

The above examples display how the notion of AGMs span models from the most popular paradigms in the field of generative networks. They have already been proved worth studying for their qualities within representation learning, semi-supervised learning and manifold learning. While the connection made in \cref{agm-definition} to atlas estimation is, at least for some of the models, not new, assembling them into a class of models makes it possible to develop geometric procedures and non-linear statistics concisely, without being model specific, as we shall see exemplified next. 

\subsection{Geodesics in Atlas Generative Models}\label{theory-atlasgeodesic}
\begin{algorithm}[t]
    \caption{Graph based geodesics for atlas generative models with encoding, generalizing \cite[Alg.~1]{chen_fast_2019}. Parameters: Number of points $N\in \N$ and cutoff margin $\varepsilon \geq 0$.}
    \begin{algorithmic}
            \STATE \textbf{1. Building graph}
            \STATE Train AGM.
            \STATE Sample $N$ points $\var{X} \sim P_{\mathrm{data}}(X)$.
            \STATE Initialize empty graph $\mathcal{G}$.
            \FOR{$i = 1, \ldots, m$}
            \STATE Find $\var{X}_i = \{x \in \var{X} \mid \psi_i(x) > \varepsilon\}$.
            \STATE Encode $\var{Z}_i = F_i(\var{X}_i)$.
            \STATE Build graph $\mathcal{G}_i$ from $\var{Z}_i$ using \cite[Alg.~1,~Part~1]{chen_fast_2019} and append to $\mathcal{G}$.
            \FOR{j = 1, \ldots, i}
            \FOR{$x$ in $\var{X}_i \cap \var{X}_j$} 
            \STATE Add edge between $F_i(x)$ and $F_j(x)$ in $\mathcal{G}$ with weight $\norm{G_i \circ F_i(x) - G_j \circ F_j(x)}_2$.
            \ENDFOR 
            \ENDFOR
            \ENDFOR
            \RETURN $\mathcal{G}$
            \vspace{10pt}
            \STATE \textbf{2. Path search}
            \STATE Same as \cite[Alg.~1,~Part~2]{chen_fast_2019}.
    \end{algorithmic}\label{alg:atlas-geodesic}
\end{algorithm}
We shall now proceed to consider geodesic paths in AGMs. It is here the significance of having a partition of unity in addition to the atlas charts comes into play. In differential geometry, a partition of unity subordinate to some open cover can be a convenience, and even sometimes necessary, as it makes explicit the weight of each covering set in any given point. The same convenience is offered by a partition of unity $\psi : \datspace \to \Delta^{m-1}$ for an AGM. In particular, we may specify areas of $\datspace$ for which a given chart is represented: Choosing some $\varepsilon \geq 0$, we may consider the open sets $U_y = \{ x\in \datspace \mid \psi_y(x) > \varepsilon\}$. Furthermore, if $\varepsilon < \frac{1}{m}$, we are guaranteed that $\bigcup_{y=1}^m U_y = \datspace$. 

The utility of the sets $\{U_y\}_{y=1}^m$ is that we may start to identify points in different coordinate spaces, on the basis of whether their image under $G$, or pre-image under $F$, is contained in intersecting areas $\cap_{y \in \sigma} U_y$ for some $\sigma \subset \mathcal{Y}$, eventually enabling us to deal with the ambiguity posed by having overlapping charts. 

Let us consider this in practice as we generalize the algorithm for graph based geodesics from \cite{chen_fast_2019}, which we briefly described in \cref{riemannian-geometry}. Suppose we have an AGM and a set of data points $\var{X} = \{x^{(1)}, \ldots ,x^{(N)}\}$. A latent graph is formed by using encodings $(z_{i,y}, y) \in \latspace \times \mathcal{Y}$ whenever $x^{(i)} \in U_y$, where $z_{i,y} = F_{y}(x^{(i)})$. In other words, for any chart $y$, for which $x^{(i)}$ is assigned the label $y$ with probability larger than $\varepsilon$, we encode $x^{(i)}$ to a latent point in this chart. Within each copy of $\latspace$, we may form a graph exactly as described in \cref{riemannian-geometry}. Each of these disjoint graphs may then be connected through edges between nodes $(z_{i,y},y)$ and $(z_{i,y'}, y')$, i.e. connecting nodes that are encodings of the same point $x^{(i)}$. 

For edges within a given chart, we shall, inspired by \cite{arvanitidis_latent_2018}, numerically approximate \cref{curvelength} with 
\begin{equation}
    \hat{L}(\gamma) = \sum_{i=0}^{n-1} \norm{G(\gamma(t_i)) - G(\gamma(t_{i+1}))}_2
    \label{curve-length-approx}
\end{equation}
where $t_i = \frac{i}{n}$ for $i = 0, \ldots, n$ for some number of steps $n\in \N$. Which weights to assign to edges connecting charts is, however, not obvious. While the geometric interpretation is that $(z_{i,y},y)$ and $(z_{i,y'}, y')$ are just different coordinates for the same point on the manifold structure and thus should have a $0$-weighted edge between them, the reality is that the decodings $G(z,y)$ and $G(z', y')$ might differ slightly. As the intuition behind the geodesic distance is that it represents the curve length along the immersed manifold structure in the surrounding space $\datspace$, another natural choice of weight would be the Euclidean distance $\norm{G_y(z_{i,y}) - G_{y'}(z_{i,y'})}_2$ in $\datspace$. As such, the weight represents the actual jump made in $\datspace$ in order to change charts. The graph building procedure is summarized in \Cref{alg:atlas-geodesic}. 

    The geodesic interpolation between latent points $(z_0, y_0)$ and $(z_1, y_1)$ may, just like in \cite{chen_fast_2019}, be found by adding each of the points to the graph, connecting them to their $k$ nearest neighbours in their respective charts, and finding the shortest path between them using the A* search algorithm. 

    In our approach to graph based geodesic paths, we have stayed as close to the algorithm presented in \cite{chen_fast_2019} as possible, including the use of encoded data to build the graph. We note, however, that this is not a canonical choice, and other heuristics may be used to create the latent space graph, e.g. sampling points using the prior $(z,y) \sim P(Z,Y)$ and connecting them to their encodings in other charts, whenever the decodings satisfy $G_y(z) \in U_{y'}$ for some $y' \neq y$. It might also be possible to find overlapping points in different coordinate charts entirely without using encoding maps, which would enable this type of geodesic path estimation for semi-AGMs as well.  

\section{Experiments}\label{experiments}
\begin{table}
    \centering
    \resizebox{0.8\textwidth}{!}{%
    \begin{tabular}{|c|c|c|c|c|c|}
        \hline
        & WAE-GAN   & \multicolumn{4}{c|}{AWAE-GAN} \\ \hline
        No. charts & - & 2 & 4 & 8 & 16 \\ \hline
        Continuous dimension & 8 & 7 & 7 & 7 & 7 \\ \hline
        Trainable parameters & 21,967,881 & 21,626,265 & 20,885,793 & 21,961,217 & 21,623,721\\ \hline
        Reconstruction error & 0.0142 / 0.0144 & 0.0134 / 0.0133 & 0.0129 / 0.0141 & 0.0126 / 0.0148 & 0.0129 / 0.0162\\ \hline
        Graph nodes & 2,000 / 2,000 & 2,020 / 2,069 & 2,175 / 2,576 & 2,546 / 2,677 & 2,887 / 3,922 \\ \hline
        Graph edges &  26.6k / 26.4k & 27.0k / 26.9k & 29.6k / 33.4k & 34.7k / 35.0k & 39.8k / 52.2k \\ \hline
        Graph diameter & 6 / 6 & 8 / 8 & 8 / 9 & 8 / 9 & 8 / 9 \\
        \hline 
\end{tabular}}
\caption{Model information (MNIST/FashionMNIST). Total trainable parameter count for encoding and decoding networks together. Reconstruction error calculated as the mean over 10.000 previously unseen data points. Number of nodes and edges, as well as diameters, are from the graphs used to approximate geodesics.}
    \label{fig:models-info}
\end{table}

\begin{figure}[t]
    \begin{subfigure}{0.65\textwidth}
    \vspace{3pt}
    \includegraphics[scale=0.25]{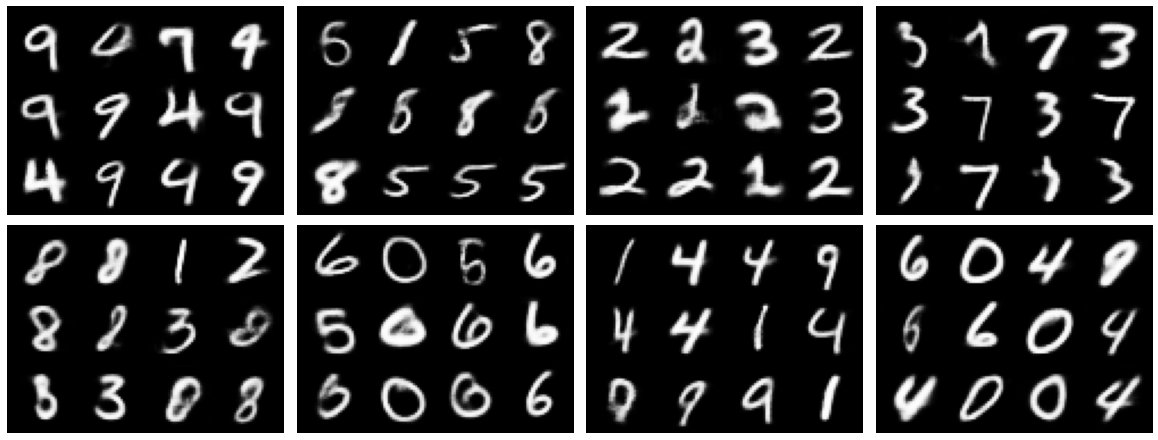}
    \includegraphics[scale=0.217]{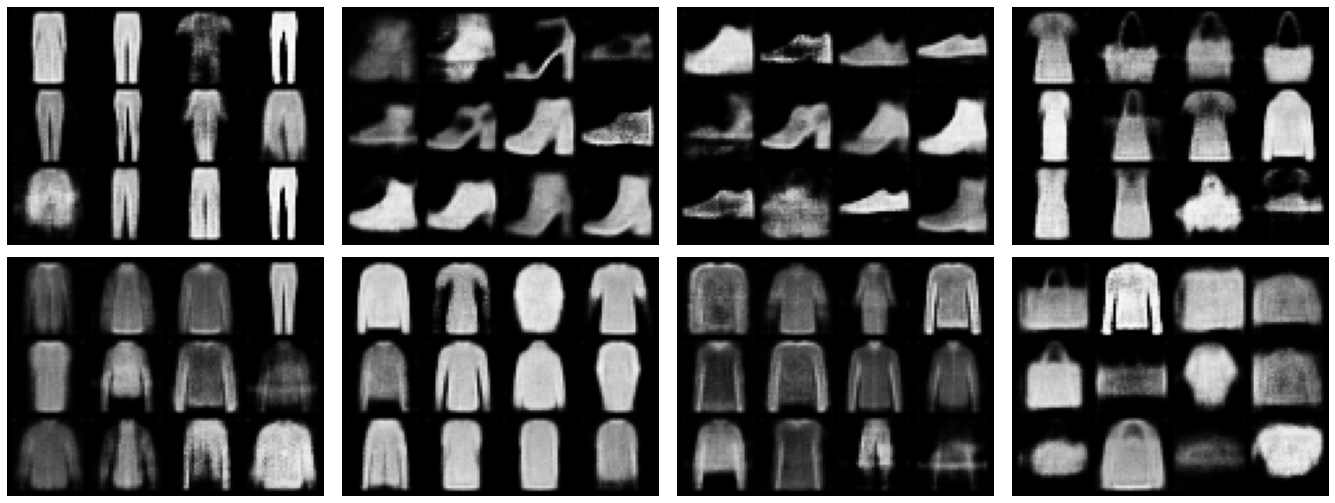}
    \vspace{5pt}
    \caption{Samples produced by the AWAE-GAN with 8 charts trained on \\MNIST (top) and FashionMNIST (bottom).}\label{fig:samples1}
\end{subfigure}%
\begin{subfigure}{0.35\textwidth}
    \centering\includegraphics[scale=0.32]{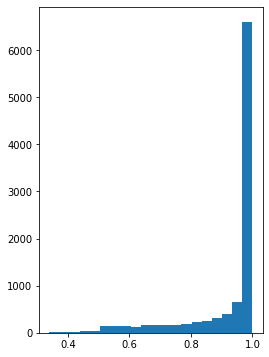} \\
    \centering\includegraphics[scale=0.32]{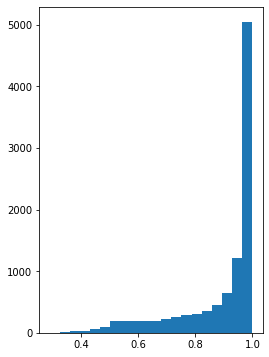}
    \caption{Chart assignment confidence for MNIST (top) and FashionMNIST (bottom).}\label{fig:samples2}
\end{subfigure}
\caption{AWAE-GAN with 8 charts. (a) Each box displays 12 random samples from a single chart. The charts approximate different parts of the data distribution with some overlaps. This is indicated by the fact that each digit/item is typically present in only 2-3 charts. (b) Histogram over chart assignment confidence $\max_y \psi_y(x)$ for 10.000 test data points in $\datspace$. Data samples are most typically assigned to a single chart with high confidence, also indicating that each chart is responsible for approximating certain parts of the data distribution.}\label{fig:samples}
\end{figure}

\begin{figure}[h]
    \centering
    \begin{subfigure}{0.3\textwidth}
        \centering
    \includegraphics[scale=0.5]{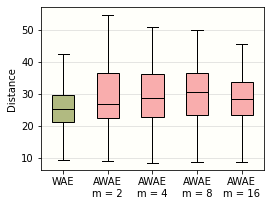}
    \includegraphics[scale=0.5]{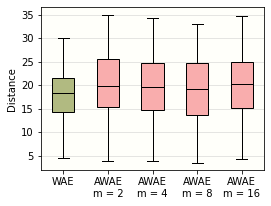}
    \caption{Geodesic distances for MNIST (top) and FashionMNIST (bottom).}\label{fig:interpolation1}
    \end{subfigure}%
    \begin{subfigure}{0.7\textwidth}
        \centering \vspace{5pt}
    \includegraphics[scale=0.22]{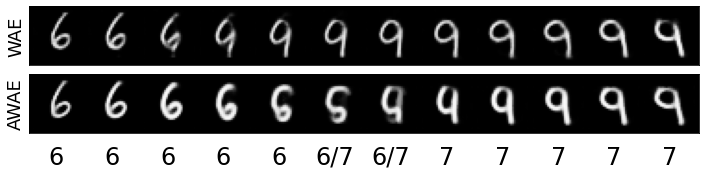}
    \includegraphics[scale=0.22]{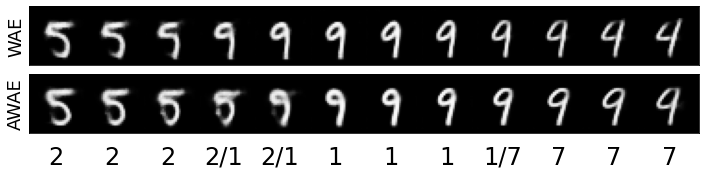}\\ \vspace{10pt}
    \includegraphics[scale=0.22]{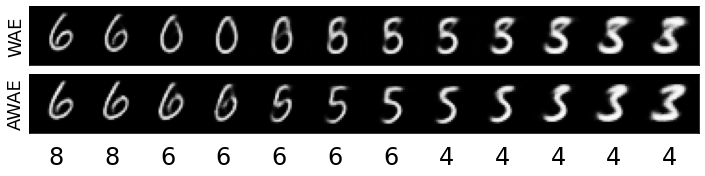}
    \includegraphics[scale=0.22]{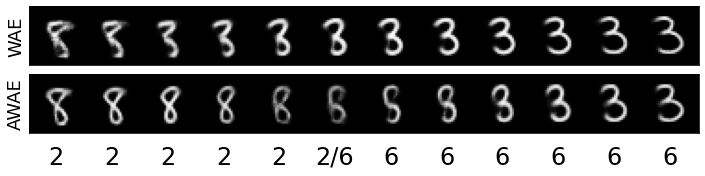}\\ \vspace{15pt}
    \includegraphics[scale=0.178]{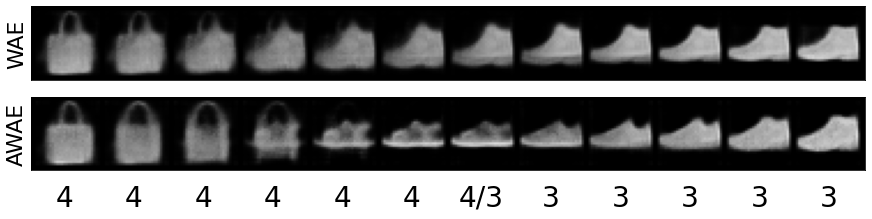}
    \includegraphics[scale=0.178]{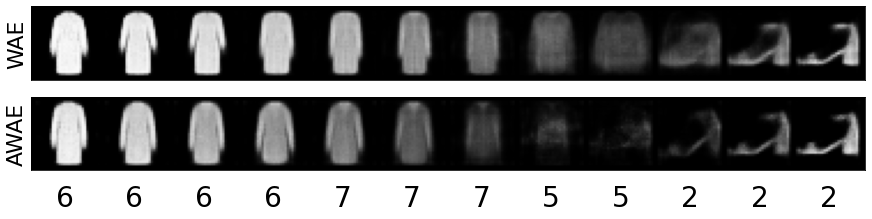}\\ \vspace{10pt}
    \includegraphics[scale=0.178]{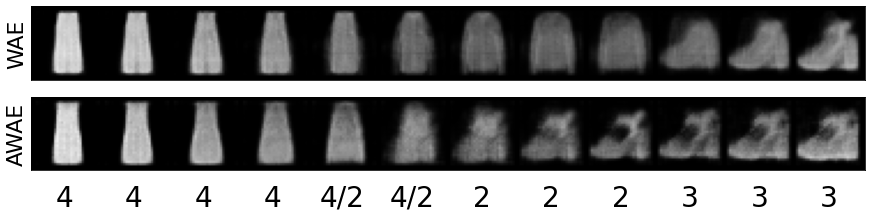}
    \includegraphics[scale=0.178]{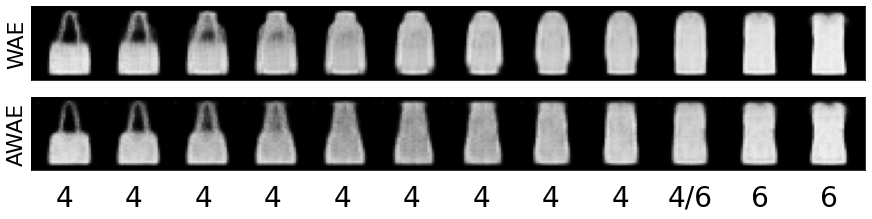}\\ \vspace{10pt}
    \caption{Geodesic interpolations for MNIST (top) and FashionMNIST (bottom).}\label{fig:interpolation2}
    \end{subfigure}
    \caption{Graph based geodesic interpolation. 
    (a) Distribution of distances for 100 graph based geodesic interpolations between the encodings of randomly sampled points $x_0, x_1 \sim P_{\mathrm{data}}(X)$. 
(b) Graph based geodesic interpolation in the WAE-GAN model (top rows) and the AWAE-GAN model with 8 charts (bottom rows). Start and end points are encodings of random test data and intermediate steps are equidistant samples along the paths with respect to the Riemannian metric. The current chart is indicated beneath each AWAE interpolation step, with '$y_1$/$y_2$' indicating an edge between charts. }\label{fig:interpolation}
\end{figure}

We have suggested how the notion of AGMs makes it possible to make sense of continuous geodesic interpolations in the latent space, even when this is hybrid discrete-continuous. In this section we experimentally verify this idea. To do so we have implemented an Atlas WAE-GAN (AWAE-GAN) \cite{korman_autoencoding_2018}, which we use to evaluate the procedure for graph based geodesics described in \cref{theory-atlasgeodesic} on the MNIST \cite{deng_mnist_2012} and the FashionMNIST datasets \cite{xiao_fashion-mnist_2017}. For comparison, we also implemented the (single chart) graph based geodesic interpolation from \cite{chen_fast_2019} on a regular WAE-GAN.

The WAE-GAN is implemented exactly as in \cite{tolstikhin_wasserstein_2018} with latent space $\R^8$ and multivariate standard Gaussian prior. For the AWAE-GAN, we replace one continuous dimension with the discrete set $\mathcal{Y}$, the size of which we vary (see \cref{fig:models-info}). The prior on the continuous part is also a multivariate Gaussian, and we use a uniform categorical distribution for $P(Y)$. The chart encoders and decoders, as well as the discrete inference network, are implemented similarly to \cite{tolstikhin_wasserstein_2018}, though to decrease model size, we use shared layers between the individual encoding charts and the partition of unity, as well as for the decoding charts. Layer sizes of the AWAE-GAN model are decreased depending on the number of charts, so that the overall model size matches that of the WAE-GAN in terms of trainable parameters (see \cref{fig:models-info}). A detailed model description can be found in \ref{appendix-model}.

The performances of the models in terms of test reconstruction errors are similar, though we observe in this experiment better performance by some of the AWAE-GANs compared to the WAE-GAN (see \cref{fig:models-info}), despite the discrete latent dimension theoretically enabling less expressiveness than a continuous dimension. This could be due to the topological representation issues discribed in \cite{schonsheck_chart_2020}, although drawing that conclusion would require more thorough investigation. 

In \cref{fig:samples1} we display random samples produced by the different charts of our AWAE-GAN with 8 charts. We observe that the same digits typically appear in only 2-3 charts, indicating that the different charts indeed represent different parts of the data distribution, with some overlaps as expected. We note further that even without regularization for improving disentanglement, such as MI-regularization, the chart assignment confidence of the model, measured as the probability of the most significant chart $\max_y \psi_y(x)$ on a test set of 10.000 previously unseen data points, is close to $1$ for a large proportion of the data, see \cref{fig:samples2}. From an atlas estimation point of view, this is a good quality, as most data is thus only assigned to one chart, while overlaps are less significant. In its own right, it is also an interesting observation that this is indeed the representation the model converges towards, even without adding any regularizing terms to enforce it.

For the graph based interpolation, we sample points for the latent space graph by encoding 2000 data points. These are connected to their 20 nearest neighbours and the edges are weighted with an approximation of the linear interpolation curve length obtained by \cref{curve-length-approx} using 15 intermediate steps.

To evaluate the performance of the geodesics on the Atlas WAE-GAN model, we pick 100 start and end points $\var{X}_{\mathrm{start}}, \var{X}_{\mathrm{end}} \sim P_{\mathrm{data}}(X)$ from a test data set and encode these points to their latent representation. Using \cite[Alg.~1]{chen_fast_2019} for the WAE-GAN model and \cref{alg:atlas-geodesic} for the AWAE-GAN model, we find the geodesic interpolations and the lengths of these paths. We see in \cref{fig:interpolation1}, that the interpolations on the AWAE-GAN model tend to be longer than on the WAE-GAN model. We expect this is a result of the chart overlaps creating bottlenecks in the graph, which is not present in the WAE-GAN graph, resulting in slightly longer interpolations. This is also coherent with the fact that the AWAE-GAN graphs have a bigger diameter than the WAE-GAN graph (see \cref{fig:models-info}) and the fact that having more charts, thus increasing the total of chart connecting edges, seems to counter the effect.  

Overall, we observe that the graph based geodesics on the AWAE-GAN does produce path lengths and interpolation quality (see \cref{fig:interpolation2}) comparable to that of the graph based geodesics on the WAE-GAN. In particular, we see in \cref{fig:interpolation2}, that this is despite intermediate transitions between charts, which we find noteworthy, given the discontinuous nature of the AGM latent space.

\section{Conclusion}
We have introduced the notion of AGMs, a class of generative networks which justifiably resemble an estimation of an atlas on the underlying manifold structure of a given data distribution $P_{\mathrm{data}}(X)$. Examples of AGMs have been surveyed, spanning different popular paradigms of generative models.

Though the non-linear nature of AGM latent spaces inherently exclude a notion of linear interpolation and Euclidean distance, we have expanded on the atlas interpretation from differntial geometry, and instead made sense of geodesic interpolations and Riemannian distance for this class of models. We have verified this in practice by presenting an algorithm generalizing the graph based approach to geodesic interpolation of \cite{chen_fast_2019}, and obtained interpolations of comparable quality to that of a non-atlas model. While the geodesic paths produced in the AGM had a tendency to be slightly longer than those of the non-atlas model, we still conclude that the suggested procedure represents a viable and novel concept of interpolation for the class of AGMs. 

Future work could include investigating the generalization of other tools from non-linear latent space analysis and statistics to the setting of AGMs. Another possible direction for further research is to improve the geometric features of the AGMs, e.g. by developing a regularizing term used during training, which produces more coherent and smooth chart overlaps in $\datspace$.

\section*{Acknowledgment}
The work presented in this paper was supported by the Villum Foundation grant 00022924 and the Novo Nordisk Foundation grant NNF18OC0052000. 

\appendix
\section{Model Implementation}\label{appendix-model}
For the experiment with graph based geodesic interpolation on the MNIST and FashionMNIST dataset, we implemented two generative networks, a WAE-GAN \cite{tolstikhin_wasserstein_2018} and an AWAE-GAN \cite{korman_autoencoding_2018}. In this section we briefly go through the details of their implementation. 

The WAE-GAN was implemented exactly like in \cite{tolstikhin_wasserstein_2018}, except we use Swish activation functions \cite{ramachandran_searching_2018} instead of ReLU, resulting in overall smooth transformations.

For the AWAE-GAN we used the same architecture for the individual charts, only scaled down to match the total number of parameters. The first several encoding layers were shared by all encoding maps and the partition of unity. The last few decoding layers was shared by all decoding maps. The discriminative networks had no shared layers. Using the same notation as in \cite{tolstikhin_wasserstein_2018} and marking shared layers with a $*$, the architectures can be abbreviated into the following diagrams\\

\textbf{Encoders:}
\begin{center}
\begin{tikzcd}[cramped,column sep=small, row sep=3pt,math mode=false]
    $x \in \R^{28\times28}$ \arrow[r] & Conv$_{h}$* \arrow[r] & BN* \arrow[r] & Swish* \\
    \hspace{54pt} \arrow[r] & Conv$_{2h}$* \arrow[r] & BN* \arrow[r] & Swish* \\
    \hspace{54pt} \arrow[r] & Conv$_{4h}$* \arrow[r] & BN* \arrow[r] & Swish* \\
    \hspace{54pt} \arrow[r] & Conv$_{8h}$ \arrow[r] & BN \arrow[r] & Swish \arrow[r] & FC$_7$
\end{tikzcd}
\end{center}

\textbf{Discrete inference (partition of unity):}
\begin{center}
\begin{tikzcd}[cramped,column sep=small, row sep=3pt,math mode=false]
    $x \in \R^{28\times28}$ \arrow[r] & Conv$_{h}$* \arrow[r] & BN* \arrow[r] & Swish* \\
    \hspace{54pt} \arrow[r] & Conv$_{2h}$* \arrow[r] & BN* \arrow[r] & Swish* \\
    \hspace{54pt} \arrow[r] & Conv$_{4h}$* \arrow[r] & BN* \arrow[r] & Swish* \\
    \hspace{54pt} \arrow[r] & Conv$_{8h}$ \arrow[r] & BN \arrow[r] & Swish \arrow[r] & FC$_m$ \arrow[r] & Softmax
\end{tikzcd}
\end{center}

\textbf{Decoders:}
\begin{center}
\begin{tikzcd}[cramped,column sep=small, row sep=3pt,math mode=false]
    \hspace{20pt}$z \in \R^{8}$ \arrow[r] & FC$_{7\times7\times8h}$ \\
    \hspace{54pt} \arrow[r] & FSConv$_{4h}$ \arrow[r] & BN \arrow[r] & Swish \\
    \hspace{54pt} \arrow[r] & FSConv$_{2h}$* \arrow[r] & BN* \arrow[r] & Swish* \arrow[r] & FSConv$_{1}$*
\end{tikzcd}
\end{center}

\textbf{Discriminators:}
\begin{center}
\begin{tikzcd}[cramped,column sep=small, row sep=3pt,math mode=false]
    \hspace{20pt}$z \in \R^{8}$ \arrow[r] & FC$_{4h}$ \arrow[r] & BN \arrow[r] & Swish \\
    \hspace{54pt} \arrow[r] & FC$_{4h}$ \arrow[r] & BN \arrow[r] & Swish \\
    \hspace{54pt} \arrow[r] & FC$_{4h}$ \arrow[r] & BN \arrow[r] & Swish \\
    \hspace{54pt} \arrow[r] & FC$_{4h}$ \arrow[r] & BN \arrow[r] & Swish \arrow[r] & FC$_1$ \arrow[r] & Sigmoid \\
\end{tikzcd}
\end{center}

The hyper-parameter $h$ was adjusted depending on the amount of charts, in order to have comparable parameter count:
\begin{center}
    \begin{tabular}{|c|c|c|c|c|c|}
        \hline 
        No. charts & 1 (WAE) & 2 & 4 & 8 & 16 \\ \hline
        $h$ & 128 & 86 & 64 & 48 & 34 \\ \hline
    \end{tabular}
\end{center}

For both the WAE-GAN and the AWAE-GAN, the hyper-parameter $\lambda$ from \cref{ot-approx} was set to 10. For the MNIST dataset we trained using mini-batches of size 100 and the ADAM optimizer with $\beta_1 = 0.5$ and $\beta_2 = 0.999$. The learning rate was initialized as $0.0002$ for the encoding (including the partition of unity) and decoding networks, and as $0.0001$ for the discriminative network, and was then halved after $5$, $25$ and $50$ epochs. For FashionMNIST mini-batches of size 128 was used and leaning rate was initialized at $0.0001$ (and $0.00005$ for the discriminative network), halving after 15, 30, 50 and 70 epochs. In both cases training was stopped after 80 epochs. 


\bibliographystyle{elsarticle-num} 
\bibliography{references.bib}

\begin{thebibliography}{10}
\expandafter\ifx\csname url\endcsname\relax
  \def\url#1{\texttt{#1}}\fi
\expandafter\ifx\csname urlprefix\endcsname\relax\def\urlprefix{URL }\fi
\expandafter\ifx\csname href\endcsname\relax
  \def\href#1#2{#2} \def\path#1{#1}\fi

\bibitem{shao_riemannian_2018}
H.~Shao, A.~Kumar, P.~T. Fletcher, The {Riemannian} {Geometry} of {Deep}
  {Generative} {Models}, in: {IEEE} {Conference} on {Computer} {Vision} and
  {Pattern} {Recognition} {Workshops} ({CVPR} {Workshops}), Computer Vision
  Foundation / {IEEE} Computer Society, 2018, pp. 315--323.

\bibitem{chen_metrics_2018}
N.~Chen, A.~Klushyn, R.~Kurle, X.~Jiang, J.~Bayer, P.~Smagt, Metrics for {Deep}
  {Generative} {Models}, in: Proceedings of the {Twenty}-{First}
  {International} {Conference} on {Artificial} {Intelligence} and {Statistics}
  ({AISTATS}), Proceedings of {Machine} {Learning} {Research}, 2018, pp.
  1540--1550.

\bibitem{arvanitidis_latent_2018}
G.~Arvanitidis, L.~K. Hansen, S.~Hauberg, Latent {Space} {Oddity}: on the
  {Curvature} of {Deep} {Generative} {Models}, in: 6th International Conference
  on Learning Representations ({ICLR}), OpenReview.net, 2018.

\bibitem{schonsheck_chart_2020}
S.~Schonsheck, J.~Chen, R.~Lai, Chart {Auto}-{Encoders} for {Manifold}
  {Structured} {Data}, CoRR\href {http://arxiv.org/abs/1912.10094}
  {\path{arXiv:1912.10094}}.

\bibitem{chen_fast_2019}
N.~Chen, F.~Ferroni, A.~Klushyn, A.~Paraschos, J.~Bayer, P.~van~der Smagt, Fast
  {Approximate} {Geodesics} for {Deep} {Generative} {Models}, in: 28th
  International Conference on Artificial Neural Networks (ICANN), Springer,
  2019, pp. 554--566.

\bibitem{deng_mnist_2012}
L.~Deng, The {MNIST} {Database} of {Handwritten} {Digit} {Images} for {Machine}
  {Learning} {Research}, IEEE Signal Processing Magazine 29 (2012) 141--142.

\bibitem{kingma_auto-encoding_2014}
D.~P. Kingma, M.~Welling, Auto-encoding variational bayes, in: 2nd
  International Conference on Learning Representations ({ICLR}),
  OpenReview.net, 2014.

\bibitem{kingma_semi-supervised_2014}
D.~P. Kingma, S.~Mohamed, D.~Jimenez~Rezende, M.~Welling, Semi-supervised
  {Learning} with {Deep} {Generative} {Models}, in: Advances in {Neural}
  {Information} {Processing} {Systems} (NIPS), Vol.~27, Curran Associates,
  Inc., 2014.

\bibitem{pineau_infocatvae_2018}
E.~Pineau, M.~Lelarge, {InfoCatVAE}: {Representation} {Learning} with
  {Categorical} {Variational} {Autoencoders}, CoRR\href
  {http://arxiv.org/abs/1806.08240} {\path{arXiv:1806.08240}}.

\bibitem{chen_infogan_2016}
X.~Chen, Y.~Duan, R.~Houthooft, J.~Schulman, I.~Sutskever, P.~Abbeel,
  {InfoGAN}: {Interpretable} {Representation} {Learning} by {Information}
  {Maximizing} {Generative} {Adversarial} {Nets}, in: Advances in {Neural}
  {Information} {Processing} {Systems} (NIPS), Vol.~29, Curran Associates,
  Inc., 2016.

\bibitem{goodfellow_generative_2014}
I.~Goodfellow, J.~Pouget-Abadie, M.~Mirza, B.~Xu, D.~Warde-Farley, S.~Ozair,
  A.~Courville, Y.~Bengio, Generative {Adversarial} {Nets}, in: Advances in
  {Neural} {Information} {Processing} {Systems} (NIPS), Vol.~27, Curran
  Associates, Inc., 2014.

\bibitem{spurr_guiding_2017}
A.~Spurr, E.~Aksan, O.~Hilliges, Guiding {InfoGAN} with {Semi}-supervision, in:
  Machine {Learning} and {Knowledge} {Discovery} in {Databases}, Lecture
  {Notes} in {Computer} {Science}, Springer International Publishing, 2017, pp.
  119--134.

\bibitem{tolstikhin_wasserstein_2018}
I.~O. Tolstikhin, O.~Bousquet, S.~Gelly, B.~Sch{\"{o}}lkopf, Wasserstein
  {Auto}-{Encoders}, in: 6th International Conference on Learning
  Representations ({ICLR}), OpenReview.net, 2018.

\bibitem{makhzani_adversarial_2016}
A.~Makhzani, J.~Shlens, N.~Jaitly, I.~Goodfellow, Adversarial {Autoencoders},
  in: 4th International Conference on Learning Representations Workshops
  ({ICLR} {Workshops}), OpenReview.net, 2016.

\bibitem{korman_autoencoding_2018}
E.~O. Korman, Autoencoding {Topology}, CoRR\href
  {http://arxiv.org/abs/1803.00156} {\path{arXiv:1803.00156}}.

\bibitem{kuhnel_latent_2018}
L.~K{\"{u}}hnel, T.~Fletcher, S.~C. Joshi, S.~Sommer, Latent {Space}
  {Non}-linear {Statistics}, CoRR\href {http://arxiv.org/abs/1805.07632}
  {\path{arXiv:1805.07632}}.

\bibitem{xiao_fashion-mnist_2017}
H.~Xiao, K.~Rasul, R.~Vollgraf, Fashion-{MNIST}: a {Novel} {Image} {Dataset}
  for {Benchmarking} {Machine} {Learning} {Algorithms}, CoRR\href
  {http://arxiv.org/abs/1708.07747} {\path{arXiv:1708.07747}}.

\bibitem{ramachandran_searching_2018}
P.~Ramachandran, B.~Zoph, Q.~V. Le, Searching for activation functions, in: 6th
  International Conference on Learning Representations Workshop ({ICLR}
  {Workshop}), OpenReview.net, 2018.

\end{thebibliography}

\end{document}